\documentclass[10pt,twocolumn,letterpaper]{article}

\usepackage[pagenumbers]{cvpr} 

\usepackage{config}

\usepackage[capitalize]{cleveref}
\crefname{section}{Sec.}{Secs.}
\Crefname{section}{Section}{Sections}
\Crefname{table}{Table}{Tables}
\crefname{table}{Tab.}{Tabs.}

\def\cvprPaperID{11277} 
\def\confName{CVPR}
\def\confYear{2022}

\begin{document}

\title{Improving Robustness with Image Filtering}

\author{Matteo Terzi, Mattia Carletti, Gian Antonio Susto\\
University of Padova\\
{\tt\small \{matteo.terzi,gianantonio.susto\}@unipd.it, mattia.carletti@studenti.unipd.it}
}
\maketitle

\begin{abstract}
Adversarial robustness is one of the most challenging problems in Deep Learning and Computer Vision research. All the state-of-the-art techniques require a time-consuming procedure that creates cleverly perturbed images. 
Due to its cost, many solutions have been proposed to avoid Adversarial Training. However, all these attempts proved ineffective as the attacker manages to exploit spurious correlations among pixels to trigger brittle features implicitly learned by the model. This paper first introduces a new image filtering scheme called Image-Graph Extractor (IGE) that extracts the fundamental nodes of an image and their connections through a graph structure. By leveraging the IGE representation, we build a new defense method, Filtering As a Defense, that does not allow the attacker to entangle pixels to create malicious patterns. Moreover, we show that data augmentation with filtered images effectively improves the model's robustness to data corruption.
We validate our techniques on CIFAR-10, CIFAR-100, and ImageNet.
\end{abstract}

\section{Introduction}
\label{sec:intro}

The present work was inspired by visual inspection of many adversarial example images.
\begin{figure}[htb]
    \centering
    \includegraphics[width=0.6\linewidth]{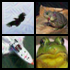}
    \caption{Micro-patterns created by $\ell_2$-bounded adversarial attacks on \ct~images.}
    \label{fig:adv_ex}
\end{figure}
As can be appreciated in~\Cref{fig:adv_ex}, where adversarial examples related to \ct~are reported, the attack seeks to create spurious micro-patterns that, although perceptually negligible to humans, dramatically impact the behavior of Deep Neural Networks (DNNs)~\cite{szegedy2013intriguing}. These artifacts give rise to many tailored `virtual edges', which leverage structural properties of the DNN architecture (specifically, the sum-aggregation operation in convolutional layers and the increasing receptive field size in deeper layers) to excite specific internal model biases. As a result, the model predictions can be easily subverted, causing input samples to be misclassified with high confidence.

Indeed, current DNNs are strongly dependent on a fragile balance of features that, if broken, can lead to a significant drop in generalization power~\cite{tsipras2018robustness,ilyas2019adversarial}.

The fundamental strength of the micro-patterns introduced by adversarial attacks lies in their simultaneous local and non-local nature.
Locality stems from the fact that the perturbation can be very precise in spatial location and directly impacts only strictly neighboring pixels. Non-locality, instead, emerges from the possibility to leverage the correlation among spatially distant pixels.
In view of this, the attacker is allowed to be very precise in choosing the optimal (relative) spatial locations so that the malign effects of perturbations can conveniently accumulate across layers.

In recent years, many defenses against adversarial examples have been proposed. However, the vast majority of the proposed approaches proved to be ineffective under stronger attacks or adaptive ones, as their false sense of robustness turned out to be related to different forms of obfuscated gradients~\cite{athalye2018obfuscated, tramer2020adaptive}.

The most reliable method to truly enforce robustness in DNNs is represented by Adversarial Training (AT)~\cite{madry2017towards}, which consists in training DNNs on adversarial samples rather than natural ones. In these settings, the training process is induced to find a configuration of parameters so that the model is robust to adversarial examples.
The spurious textures characterizing adversarial examples let the model infer that a peculiar configuration of a few distant pixels is, in principle, not correlated with the task or that a particular local texture is not predictive. As this process takes place during training, the model can adapt its parameters accordingly.

As mentioned above, many issues related to the correlation of spatially distant pixels are due to the model architecture itself rather than the training procedure or the training data. Therefore, major enhancements in adversarial robustness should come from novel architecture layouts taking into account the structural weaknesses discussed above. However, significant improvements can still be achieved by acting on the input space.

Based on the above intuitions, this paper presents a novel image filtering framework aimed at improving robustness against adversarial attacks without resorting to AT. While it is not the first contribution introducing techniques based on input manipulation to prevent adversarial attacks~\cite{guo2017countering, lu2017no, dziugaite2016study}, this work differs significantly from prior literature in the fundamental concepts at the basis of its functioning. The main difference between our method and others relying on input transformations is that we \textit{explicitly} remove textures from input images while preserving the informative features. This is achieved through an iterative filtering procedure based on simple rules that do not require additional training routines. Details and analysis of our proposed approach are provided in~\Cref{sec:ige} and~\Cref{sec:fad}.

The image filtering strategy proposed in this paper can be exploited as a data augmentation, which we show being beneficial also for robustness against common corruptions (e.g., shot noise, pixelate, glass blur). Additionally, our framework allows for high flexibility in the choice of its key components, as long as its underpinning principles are not violated. 

Our contributions can be summarized as follows:
\begin{itemize}
    \item We introduce a novel image filtering framework, dubbed Image-Graph Extractor (IGE), whose main objective is to remove textures while preserving the image's semantic content. Our framework is based on an implicit graph representation of images, the Image-Graph (IG).
    \item We propose an effective defense against bounded adversarial attacks, called Filtering As a Defense (FAD), leveraging the textures suppression effect of IGE.
    \item We exploit filtered images produced through the IGE filtering framework as a data augmentation strategy. We show that models trained with our data augmentation strategy are more robust against common corruptions such as shot noise, pixelate, glass blur.
\end{itemize}

While IGE has been primarily conceived and designed to enforce robustness through explicit textures suppression in the input image, the intermediate graph representation necessary for the iterative filtering procedure might prove useful in many other applications. In fact, the IG associated with the input image enables the emergence of structure in an unstructured domain. The extracted structure, if adequately exploited, might boost performance in downstream tasks. This will be the subject of further investigations in future research.

The paper is organized as follows. In~\Cref{sec:related_work} we review existing literature related to our work. \Cref{sec:notation} lays out the necessary notation and preliminary concepts. In~\Cref{sec:ige} we introduce and analyze our IGE framework, while~\Cref{sec:fad} describes the FAD defense. In~\Cref{sec:results} we present the results of our experiments. In~\Cref{sec:discussion} we discuss limitations and potential improvements of our framework. In~\Cref{sec:conclusions} we summarize our findings and point to some possible research directions for future work.

\section{Related Work}
\label{sec:related_work}
Since the discovery of the existence of adversarial examples~\cite{szegedy2013intriguing}, countless attempts to make image classifiers robust to adversarial perturbations have appeared in the literature. As anticipated in~\Cref{sec:intro}, current state-of-the-art methods to enforce robustness directly leverage adversarial examples during training~\cite{goodfellow2014explaining, kurakin2016adversarial, kannan2018adversarial, madry2017towards}. A major limitation affecting AT is the high computational cost as adversarial examples are typically crafted through iterative optimization routines. For this reason, significant efforts have been made to improve the robustness of models without resorting to AT. Proposed solutions cover a wide range of techniques based on curvature regularization~\cite{Moosavi-Dezfooli:2018aa}, robust optimization to improve local stability~\cite{shaham2015understanding}, the use of additional unlabeled data~\cite{carmon2019unlabeled}, local linearization~\cite{qin2019adversarial}, Parseval networks~\cite{cisse2017parseval}, defensive distillation~\cite{papernot2016distillation}, model ensembles~\cite{pang2019improving}, channel-wise activations suppressing~\cite{bai2021improving}, feature denoising~\cite{xie2019feature}, self-supervised learning for adversarial purification~\cite{shi2020online}, and input manipulations~\cite{guo2017countering, dziugaite2016study, lu2017no}. All the listed techniques, except those based on input manipulations, require training the model or an auxiliary module from scratch. Our method, instead, can be used in combination with pretrained models.

In~\cite{lu2017no} the authors analyze the effect of image rescaling on adversarial examples. \cite{dziugaite2016study} explores the possibility to improve robustness through JPG (re)compression, based on the intuition that adversarial perturbations are unlikely to leave an image in the space of JPG images. In~\cite{guo2017countering}, the authors assess the effectiveness of input transformations based on image cropping and rescaling, bit-depth reduction, JPEG compression, total variance minimization, and image quilting. The main objective is to remove the adversarial perturbations from images, while preserving sufficient information to correctly classify them. Unlike the mentioned input-based  techniques that only implicitly (or not at all) induce mitigation of the bias towards textures, our method makes this effect explicit by directly removing textures from the input image.  

\section{Preliminaries and Notation}
\label{sec:notation}
We introduce here the notation used throughout this paper. We focus on image classification problems where a classifier $f_{\boldsymbol{\theta}}(\cdot)$ is encoded by a DNN parameterized by $\boldsymbol{\theta} \in \mathbb{R}^{|\boldsymbol{\theta}|}$. The parameters vector $\boldsymbol{\theta}$ is optimized by minimizing a suitable loss function $\mathcal{L}(\mathbf{x}, y; \boldsymbol{\theta})$ on samples $(\mathbf{x}, y)$ drawn from the underlying training data distribution $\mathcal{D}$. A training sample consists of an RGB (normalized) image $\mathbf{x} \in \mathcal{X} \subset [0,1]^{3 \times h \times w}$ and the corresponding true label $y \in \{ 1, \dots, K \}$ (as common practice, we assume $h=w$). We consider the cross-entropy loss optimized using Stochastic Gradient Descent (SGD)~\cite{bottou2018optimization}, which updates the parameters $\boldsymbol{\theta}$ with a noisy estimate of the gradient computed from a mini-batch of samples. The predicted label corresponding to the input image $\mathbf{x}$ is denoted by $\hat{y}= f_{\boldsymbol{\theta}}(\mathbf{x})$.

\paragraph{PGD}
Given an input image $\mathbf{x}$, Projected Gradient Descent (PGD)~\cite{madry2017towards} is a gradient-based method for finding an adversarial example $\mathbf{x}'$ that satisfies the condition $\|\mathbf{x}' -\mathbf{x}\|_p \leq \epsilon$, for given norm $\| \cdot \|_{p}$ and perturbation budget $\epsilon>0$. Let $B$ denote the $\ell_p$-ball of radius $\epsilon$ centered at $\mathbf{x}$.
The attack is initialized at a random point $\mathbf{x}_0 \in B$, and repeatedly applies (for a pre-specified number of iterations) the following update rule:
\begin{align*}
\mathbf{x}_{k+1} &= \Pi_{B}(\mathbf{x}_k + \alpha \cdot \mathbf{g}) \\
\quad \text{for } \mathbf{g} &= \argmax_{\|\mathbf{u}\|_p \leq 1} \mathbf{u}^\top \nabla_{\mathbf{x}_k} \mathcal{L}(\mathbf{x}_k, y) \;.
\end{align*}
Here, $\mathcal{L}(\mathbf{x}, y)$ is a loss function (e.g., cross-entropy), $\alpha$ is a step-size, $\Pi_B$ projects an input onto $B$, and $\mathbf{g}$ is the gradient, that represents the steepest ascent direction for a given $\ell_p$-norm. 

\section{Image Graph Extractor}
\label{sec:ige}
In this Section we introduce the IGE framework. For the sake of readability, we first highlight the guiding principles at its core (\Cref{sec:guiding_principles}) and describe thoroughly its key components, i.e., the merging rule (\Cref{sec:merging_rule}) and the iterative filtering procedure (\Cref{sec:iter_filtering}). All the design choices characterizing the specific instance of the IGE method used for our experiments are discussed in more detail in~\Cref{sec:design_choices}. In order to satisfy parallel and efficient image processing we developed IGE in \texttt{CUDA/C++} and NVIDIA \texttt{Thrust}. Implementation details are given in~\Cref{sec:impl_details}.

\subsection{Guiding Principles}
\label{sec:guiding_principles}
The information contained in an image depends on the effective resolution $r$ we see at: when `zooming out' the image (i.e., decreasing $r$), pixels that are distinct at higher resolutions are no more distinguishable and merge together forming patches with larger size. Conversely, finer details emerge when  `zooming in' the image (i.e., increasing $r$) and patches split into multiple sub-patches with smaller sizes.
Based on these observations, an image $\mathbf{x}$ at a given resolution $r$ can be represented as a graph, where nodes correspond to patches of similar pixels and edges encode information about connectivity of the different patches. The size of each node amounts to the number of pixels the corresponding patch is composed of.
As a result, each image can be partitioned into a collection of patches according to an underlying graph representation.

\subsection{Merging Rule}
\label{sec:merging_rule}
The objective of IGE is to emulate the perceptual effects described in ~\Cref{sec:guiding_principles}. While the produced filtered images reveal complex elaborations that might be hard to model, IGE is driven by a simple merging rule, which we describe hereinafter.

Let $\mathbf{x}^r$ be an image at resolution $r$ and let $\mathcal{G}^r=(\mathcal{V}^r, \mathcal{E}^r)$ be the corresponding IG, where $\mathcal{V}^r$ is the set of nodes and $\mathcal{E}^r$ the set of edges at resolution $r$. For simplicity, we assume edges to be binary-valued, with $e_{i,j}^r=1$ indicating that $v_i^r$ and $v_j^r$ are neighboring nodes ($e_{i,j}^r \in \mathcal{E}^r$ and $v_i^r, v_j^r \in \mathcal{V}^r$). In the image domain, this translates into the fact that patches corresponding to $v_i^r$ and $v_j^r$ are neighbors (according to some specified criterion). Let $s_i^r$ and $s_j^r$ be the size of nodes $v_i^r$ and $v_j^r$, respectively. We define the color associated with node $v_i^r$ (dubbed $c_i^r$) as the average value of pixels in $v_i^r$. Given an opportune color distance $d_c$, we define the \textit{adjusted distance} $d_a$ as
\begin{equation}\label{eq:adj_dist}
    d_a(c_i^r,c_j^r; s_i^r, s_j^r, r) = \phi(d_c(c_i^r,c_j^r), s_i^r, s_j^r, r)
\end{equation}
for some function $\phi(\cdot)$. Notice that $d_a$ is function of the color distance $d_c(c_i^r,c_j^r)$, the size of nodes $s_i^r$ and $s_j^r$, and the resolution $r$.
Neighboring nodes $v_i^r$ and $v_j^r$ are merged if
\begin{equation}\label{eq:merge_rule}
    d_a(c_i^r, c_j^r; s_i^r, s_j^r, r) < \mergeth(d_0, r)
\end{equation}
where $\mergeth(d_0, r)$ is the threshold (function of the resolution $r$) that determines if two nodes are in the same patch at a given resolution $r$. The parameter $d_0$ represents the minimum color distance perceived at maximum resolution $r_0$, corresponding to the resolution of the original (unfiltered) image $\mathbf{x}=\mathbf{x}^{r_0}$.
We assume by convention that $r_0=1$. In order to be consistent with the guiding principles in~\Cref{sec:guiding_principles}, $\mergeth$ must be a monotonically decreasing function with respect to $r$, that is $\dfrac{\partial \mergeth(d_0, r)}{\partial r} < 0$. Indeed, as we zoom out the image ($r$ decreases), pixels that are more and more distant from each other in color space become indistinguishable, resulting in a less strict threshold.

The adjustment of the color distance $d_c$ modeled by~\Cref{eq:adj_dist} takes into account the size of nodes. The rationale of the proposed procedure can be better explained by an example: a black 1-pixel node surrounded by white pixels becomes perceptually indistinguishable from its neighbors at a far larger resolution if compared to a black 5-pixels node.

In summary, the core of the IGE framework is represented by the merging rule described in this Section. The key ingredients defining the rule are: the criterion used to determine neighboring patches, the color distance $d_c$, the adjusted distance $d_a$ (specifically, the function $\phi$), the minimum color distance perceived at maximum resolution $d_0$, and the threshold function $\mergeth$. The choice of the latters is subject to the following constraints:
\begin{itemize}
    \item[C1.] If all the variables but $r$ are kept fixed, it is required that $\frac{\partial \mergeth(r)}{\partial r} < 0$ and/or $\frac{\partial d_a(r)}{\partial r} > 0$.
    \item[C2.] If all the variables but $s_i^r$ are kept fixed, it is required that $\frac{\partial d_a(s_i^r)}{\partial s_i^r} > 0$.
    \item[C3.] If all the variables but $d_c$ are kept fixed, $\displaystyle\frac{\partial d_a}{\partial d_c} > 0$.
\end{itemize}

The IGE framework is general in that it allows flexibility in the choice of the key ingredients as long as the constraints listed above are satisfied. In~\Cref{sec:design_choices} are given the specific design choices adopted in this work.

\subsection{Iterative Filtering}
\label{sec:iter_filtering}

The iterative filtering procedure consists in repeatedly performing merging operations, after a preliminary phase for initialization. The functioning of the whole IGE framework is summarized in~\Cref{alg:IGE}.

Let $\targetres$ be the \textit{target resolution} and $r_0=1$ the initial resolution corresponding to the original image $\mathbf{x}$, as mentioned in~\Cref{sec:merging_rule}. In our framework, the target resolution $\targetres \in \left] 0, r_0 \right[$, together with the \textit{filtering step size} $\dr$, defines the number of filtering steps $\niter=\frac{1-\targetres}{\dr}$ to be performed.

The initialization phase is aimed at extracting the IG $\mathcal{G}^{r_0}$ associated with the original image $\mathbf{x}=\mathbf{x}^{r_0}$. We represent this operation with the generic function $\mathtt{img2graph}$. 

The filtering routine can be seen as an anti-causal discrete dynamical system $\mathcal{G}^{r-\dr} = m(\mathcal{G}^{r})$, where $m$ is the map obtaining $\mathcal{G}^{r-\dr}$ from $\mathcal{G}^{r}$. Specifically, $m$ is composed of the following elementary operations:
\begin{enumerate}
    \item Merge nodes according to~\Cref{eq:merge_rule} (we denote this operation with the function $\mathtt{merge}$).
    \item Update the color value of each node by averaging the values of all pixels in the node (we denote this operation with the function $\mathtt{avg\_color}$).
\end{enumerate}
The map $m$ is applied iteratively $\niter$ times until the target resolution $\targetres$ is reached. The final step consists in reconstructing the filtered image at the target resolution from the final IG $\mathcal{G}^{\targetres}$. We represent this operation with the generic function $\mathtt{graph2img}$. 

A potential issue that might arise is the co-existence, for a specific node, of multiple candidate nodes for the merging operation. Although not optimal from a computational point of view, one trivial solution would be to choose the candidate node with the smallest adjusted distance. In this regard, the filtering step size $\dr$ controls the goodness of the filtering process. Small values of $\dr$ help in mitigating the problem of multiple candidate nodes for merging. Since $N_r \propto \dr^{-1}$, also this strategy should call for thoughtful considerations about computational cost. In light of this, we set $\dr$ to the biggest value that guarantees the required quality of the filtered images.

\begin{figure}[htbp]
    \centering
    \includegraphics[width=0.24\textwidth]{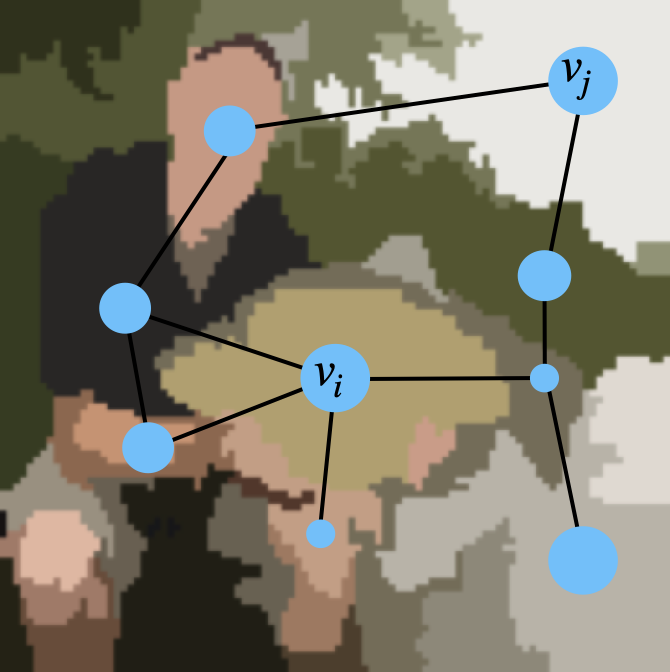}
    \caption{A simplified example of (part of) an IG associated with an image from the \imagenet~dataset. For graphical purposes, only main nodes are displayed.}
    \label{fig:graphrep}
\end{figure}

Notice that the IG associated with the filtered image at the target resolution $\targetres$ can be extracted with no extra computations as it comes as an intermediate by-product of the filtering procedure. The IG can be enriched by considering additional node attributes besides node color (e.g., shape) and encoding additional information about connectivity as edge attributes. 
A simplified example of IG is depicted in~\Cref{fig:graphrep}.

\begin{algorithm}[t]
\DontPrintSemicolon
  \KwInput{original image $\mathbf{x}$}
  \KwOutput{filtered image $\mathbf{x}^{r^*}$}
  \Param{$\targetres$, $d_c$, $d_a$, $d_0$, $\mergeth$}
  $\mathcal{G}^{r} = \mathtt{img2graph}(\mathbf{x}^{r_0})$ \tcp*{init}
  $r=r_0-\dr$ \tcp*{update res} 
  \While{$r > \targetres$}
   {
   		$\mathcal{G}^{r} = \mathtt{merge}(\mathcal{G}^r; d_c, d_a, d_0, \mergeth)$\;
   		$\mathcal{G}^{r} = \mathtt{avg\_color}(\mathcal{G}^r)$\;
   		$r=r-\dr$ \tcp*{update res} 
   }
   $\mathbf{x}^{r^*} = \mathtt{graph2img}(\mathcal{G}^r)$ \tcp*{rec filt img}
\caption{IGE pseudocode.} 
\label{alg:IGE}
\end{algorithm}

\subsection{Design Choices}
\label{sec:design_choices}
As stated in~\Cref{sec:merging_rule}, to define an instantiation of the IGE method we need to specify: (i) the criterion used to define neighboring patches (i.e., the minimum number of neighboring pixels), (ii) the color distance $d_c$, (iii) the adjusted distance $d_a$ (specifically, we should define $\phi$), (iv) the minimum color distance perceived at maximum resolution $d_0$ and (v) the merging threshold function $\mergeth$.

As for (i), we consider two distinct patches as neighbors if their connection is at least one pixel. While simple, this choice could cause merges that are not optimal. A possible improvement may come by defining a different threshold.

In respect of the color distance $d_c$, we tested two formulations: CIEDE2000~\cite{sharma2005ciede2000} and Euclidean. While CIEDE2000 offers more human-aligned color perception, we did not appreciated significant differences. Thus, we opted for the Euclidean distance. Given two RGB vectors $p=\left[ p_R, p_G, p_B \right]^{\top}$ and $q=\left[ q_R, q_G, q_B \right]^{\top}$ their Euclidean distance is given by $d_c (p,q)=\sqrt{(p_R-q_R)^2 + (p_G - q_G)^2 + (p_B - q_B)^2}$.

For the adjusted distance $d_a$, we adopted the following formulation
\begin{align}
d_a &= d_c (c_i^r,c_j^r) \cdot \tilde{\phi}(s_i^r,s_j^r,r), \\
\tilde{\phi}(s_i^r,s_j^r,r) &= {\Big(1 + e^{(-\alpha(\min\{s_i^r,s_j^r\} - \frac{\beta}{r}))}}\Big)^{-1/r}
\end{align}
where the color distance $d_c$ and adjustment $\tilde{\phi}$ terms are decoupled. We set $\alpha=0.04$ and $\beta=10$.

The term $\tilde{\phi}$ has the objective of reducing the perceived color distance when the resolution is low and/or the size of nodes is small. The size adjustment is represented by the term $\min\{s_i^r,s_j^r\}$, meaning that the decision depends on the smallest node. This is aimed to reproduce the behavior of vanishing nodes as we zoom out the image. However, this rule may be refined by also considering the relative size, that is $s_i^r/s_j^r$. The term $\frac{\beta}{r}$ is helpful to incentivize the early merge of small nodes.
We want to remark that we tested several rules $\phi$. Interestingly, the quality of results is stable with respect to the choice of $\phi$ as soon as it satisfies the constraints of the IGE framework.

Regarding the threshold function $\tau$ we investigated two different forms:
\begin{itemize}
    \item [T1]: $\tau(d_0,r) = \frac{d_0}{r}$;
     \item [T2]: $\tau(d_0,r) = d_0 \frac{1 - (r-r_{m})}{r_{m}}$, where $r_m=0.1$.
\end{itemize}
T1 is not linear. Thus, at low resolutions a small change in $r$ can have a visible impact on filtering. We opted for T2 even if also T1 provided good results. Moreover, we would like to remark that there is no `correct' rule. The form controls how we establish to perceive the zooming and image structures. We set $d_0=0.03$.


\section{Filtering As a Defense}
\label{sec:fad}
In this Section we describe FAD, our proposed defense against adversarial attacks leveraging the IGE iterative filtering procedure. We first discuss the motivation in~\Cref{sec:fad_motivation} and then introduce the full defense protocol in~\Cref{sec:fad_protocol}.

\begin{figure*}[h!]
    \centering
    \includegraphics[width=0.95\textwidth]{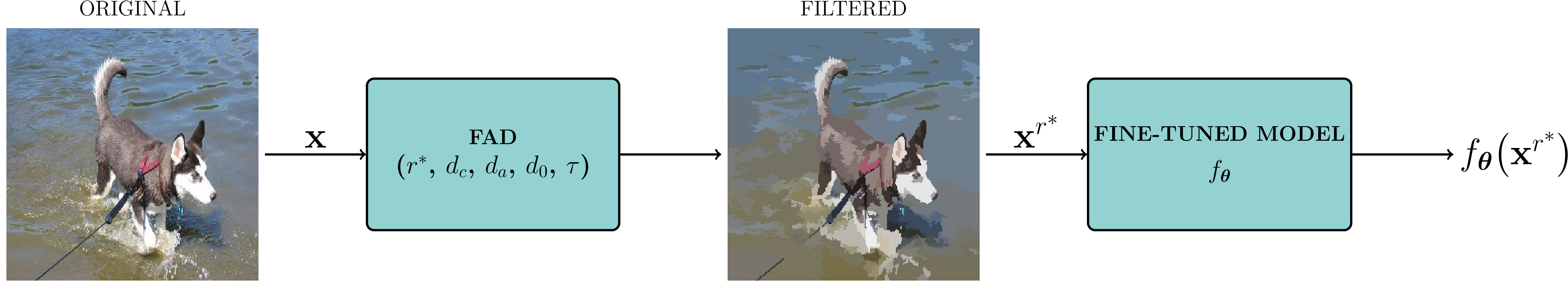}
    \caption{Block diagram describing the FAD defense.}
    \label{fig:fad}
\end{figure*}

\subsection{Motivation}
\label{sec:fad_motivation}
We want to provide an intuitive argument about the effectiveness of FAD.
Let $\mathbf{x}$ be an input image of $n = h \times w$ pixels and let $\mathbf{x}^{\targetres}$ its filtered counterpart at a given target resolution $\targetres$. We define $p$ as the \textit{effective number of pixels} in the filtered image $\mathbf{x}^{\targetres}$, that is, the number of patches the filtered image is composed of. In the context of the IGE framework, $p$ is equivalent to the number of nodes in the IG $\mathcal{G}^{\targetres}$ associated with the filtered image $\mathbf{x}^{\targetres}$.
Intuitively, we may think the filtered image as a $\sqrt{p} \times \sqrt{p}$ image. As $p \le n$ (and potentially $p \ll n$), it is evident that the total number of directions that an attacker can exploit decreases significantly if compared to the original input image.

As discussed in~\Cref{sec:intro}, in order to be effective the attacker must have the possibility to create local micro-patterns characterized by \textit{precise} relative spatial locations. Said micro-patterns, correlated together across layers, cause the classification error. In standard settings (i.e., when dealing with unfiltered images), the attacker has the freedom to perturb the image along almost any direction not too far from the original one. In visual terms, this amounts to the fact that the adversarial example is perceptually very similar to the original image. When considering filtered images, this assumption is not valid anymore and the actual number of effective directions the attacker can focus on is limited. This fact can be intuitively explained by comparing the underlying structures of original and filtered images. For original images, the structure is represented by a rigid uniform grid: if we assume $8$-connectivity, each pixel's neighborhood has fixed size equal to $8$. The effect of perturbing a single pixel directly impacts (in terms of created `virtual edges') only the 8 pixel's neighbors. For filtered images, instead, the structure is represented by a non-uniform grid defined by the IG associated with the image. This allows patches (corresponding to nodes in the IG) to potentially border with many other patches. In the worst case scenario, from the perspective of the attacker, a single patch could border with \textit{all} other patches. In this case, the perturbation of a single patch would impact all other patches (in terms of created `virtual edges'), leading to a loss in precision for the attacker.

\subsection{Defense Protocol}
\label{sec:fad_protocol}

The IGE filtering procedure outlined in~\Cref{sec:ige} causes a shift in the input image statistics that results in distribution shifts of batch norm layers' inputs. Thus, especially in complex datasets like \imagenet, models are less accurate when evaluated on filtered images as population statistics estimated on training data do not match statistics of test data.

To overcome this mismatch, the first phase of the protocol is represented by fine-tuning the pretrained model to align to filtered images. Since the semantic content of input images is preserved, the fine-tuning process requires only a few epochs to converge.

The second phase is the defense itself, which consists in running the IGE iterative filtering routine as a pre-processing step as soon as an input image is queried. \Cref{fig:fad} reports the described FAD defense. Notice that the FAD defense module inherits the parameters defining the specific instantiation of the IGE filtering method.

\section{Results}
\label{sec:results}
In this Section we present three sets of results: (i) the visual inspection of filtered images, (ii) FAD under attack, and (iii) data augmentation with filtered images.
Additional analyses are presented in the~\Cref{app:ige,app:fad}.

\subsection{Filtering}
\begin{figure*}[htbp]
\centering
\begin{subfigure}[t]{0.48\textwidth}
\includegraphics[width=1\textwidth]{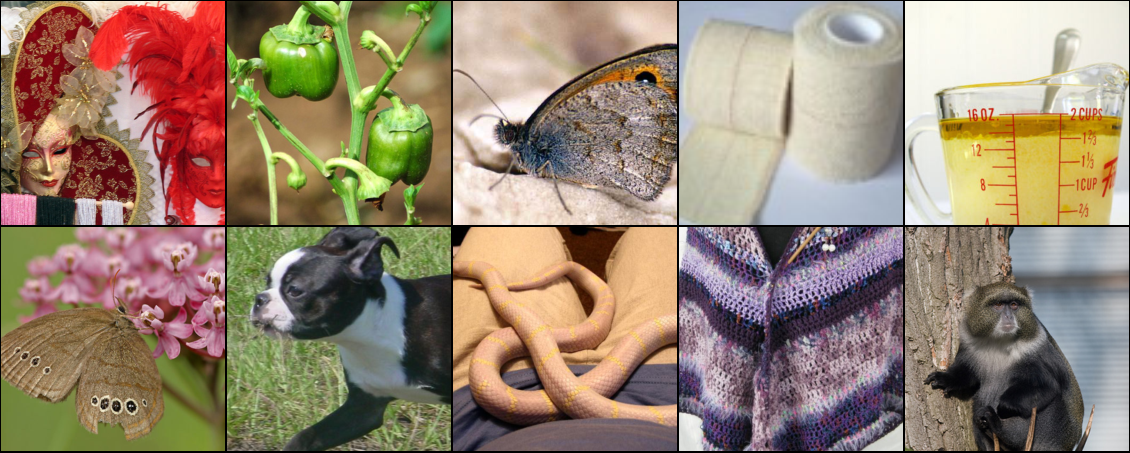}
\caption{Original Images.}
\end{subfigure}
\begin{subfigure}[t]{0.48\textwidth}
\includegraphics[width=\textwidth]{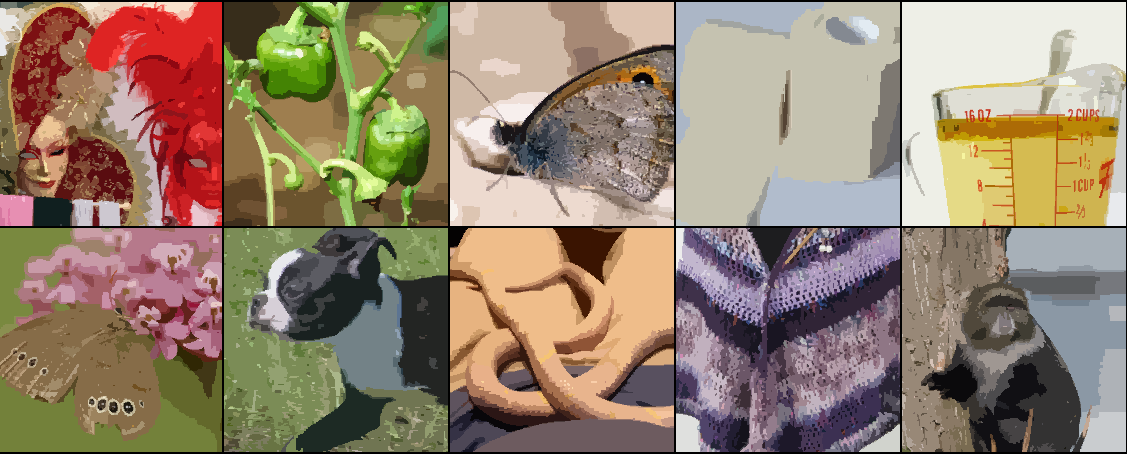}
\caption{Filtered Images.}
\end{subfigure}
\caption{Examples of filtered \imagenet~images with $\targetres=0.6$ and $\dr=0.1$.}
\label{fig:example_filtering}
\end{figure*}

In~\Cref{fig:res_ex} we show the progression of the iterative filtering process, i.e., the resulting filtered images as the target resolution varies. When reducing the target resolution, the number of nodes progressively decreases. However, the semantic content of the image remains preserved.
In~\Cref{fig:example_filtering} we show few examples of filtered images with fixed target resolution $\targetres=0.6$. The final number of nodes depends on the complexity of the original image. On average, at target resolution $\targetres=0.6$, images have roughly $[150,1000]$ nodes.

\newcommand{\fsize}{0.24}
\begin{figure*}[htbp]
    \centering
\includegraphics[width=\fsize\textwidth]{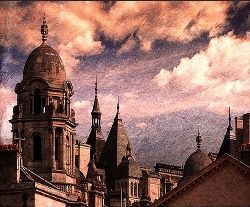}
\includegraphics[width=\fsize\textwidth]{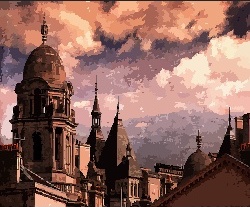}
\includegraphics[width=\fsize\textwidth]{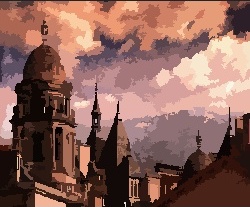}
\includegraphics[width=\fsize\textwidth]{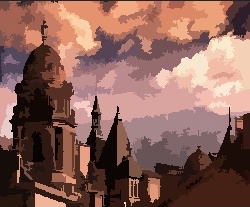}
    \caption{Example of the effect of the IGE filtering when decreasing the target resolution (from left to right).}
    \label{fig:res_ex}
\end{figure*}

\subsection{Robustness Evaluation}
As discussed by~\cite{tramer2020adaptive}, one should develop adaptive attacks to test the robustness of a defense. In this context, the Backward Pass Differentiable Approximation (BPDA) is a gradient approximation technique~\cite{athalye2018obfuscated} for dealing with non-differentiable components. 
Let $\tilde{f}$ be the composition $\tilde{f}(x) = f \circ h(x)$ where $h$ is non-differentiable (or hard to differentiate). In our case $h$ is the FAD module and $f$ represents the fine-tuned model. To approximate the total gradient $\nabla_x \tilde{f}(x)$, BPDA needs to find a differentiable function $g(x)$ such that $g(x) \approx h(x)$. Then, when computing the gradient $\nabla_x \tilde{f}(x)$, the forward pass is a standard forward pass through $\tilde{f}$ (including the forward pass through $h$), while in the backward pass $h$ is replaced by $g$.
In our case, a reasonable choice for $g$ is the identity function $g(x)=x$ as the FAD module is a simple `denoising' layer.

A vanilla application of BPDA may not be effective as one step of gradient might not allow the attacker to craft adversarial patches. 
Hence, we apply BPDA (as an outer loop) with the only difference of an inner attack (addressed to $f$) where multiple PGD steps are applied.

\subsection{Experimental Settings}
\label{sec:exp_settings}
We validate FAD with the datasets \ct~\cite{cifar}, \ch~\cite{cifar} and \imagenet~\cite{deng2009imagenet}.
As discussed in~\Cref{sec:fad_protocol}, filtered images exhibit different statistics if compared to their original counterparts. For this reason, the first phase of the FAD protocol is a fine-tuning of the model. While the optimal solution would be to fine-tune the model with the same resolution used in the FAD module, we decided to fine-tune the model for only one resolution $\targetres=0.6$. We do this as it provides a good balance between simplicity and accuracy.
For all the datasets, we employ ResNet-18~\cite{he2016identity} architectures trained for 150 epochs, initial learning rate $\eta=0.1$ and a drop of $\eta$ by a factor $10$ every 50 epochs.

As regards BPDA, we evaluate FAD with $50$ outer steps and $20$ inner steps. Attacks are bounded in $\ell_2$-norm. Due to computational constraints, we tested on $30\%$ of the test sets with shuffling.



\renewcommand{\arraystretch}{1}
\begin{table*}[htbp]
    \centering
    \resizebox{0.95\textwidth}{!}{%
    \begin{tabular}{llrrrrrcrrrrr}
\toprule
         & {} & \multicolumn{5}{c}{Accuracy} & \phantom{abc} & \multicolumn{5}{c}{Loss} \\
         &  & \multicolumn{5}{c}{\textbf{$\epsilon_{test}$}} && \multicolumn{5}{c}{\textbf{$\epsilon_{test}$}} \\
         \cmidrule{3-7} \cmidrule{9-13}
         & $\targetres$ & 0 &    1 &    2 &    3 &    4 &&      0 &   1 &   2 &    3 &    4 \\
\cmidrule{2-7} \cmidrule{9-13}
\multirow{5}{*}{\ct} & 0.4 &     78.45 &  70.15 &  55.37 &  35.09 &  21.16 &&    0.68 &  1.03 &  1.74 &   3.19 &   4.56 \\
         & 0.5 &     82.32 &  74.38 &  52.67 &  27.31 &  11.36 &&     0.54 &  0.85 &  1.91 &   4.17 &   6.48 \\
         & 0.6 &     85.06 &  73.18 &  49.80 &  18.26 &   4.39 &&    0.48 &  0.91 &  2.16 &   5.56 &   9.44 \\
         & 0.7 &     86.07 &  73.63 &  44.50 &   8.66 &   0.68 &&     0.44 &  0.91 &  2.68 &   8.14 &  13.95 \\
         & 0.8 &     89.55 &  74.48 &  33.40 &   1.82 &   0.00 &&     0.33 &  0.90 &  3.87 &  12.57 &  20.18 \\
\cmidrule{2-7} \cmidrule{9-13}
\multirow{5}{*}{\ch} & 0.4 &     41.08 &  34.99 &  25.91 &  19.04 &  15.01 &&     2.90 &  3.47 &  4.49 &   5.63 &   6.56 \\
         & 0.5 &     47.88 &  38.74 &  27.25 &  16.50 &  12.79 &&     2.47 &  3.17 &  4.59 &   6.35 &   7.58 \\
         & 0.6 &     52.60 &  41.24 &  25.52 &  13.93 &   7.49 &&     2.14 &  3.09 &  4.97 &   7.40 &   9.80 \\
         & 0.7 &     59.54 &  43.23 &  23.40 &   8.72 &   2.99 &&    1.80 &  3.03 &  5.45 &   9.54 &  13.82 \\
         & 0.8 &     63.74 &  45.31 &  19.17 &   3.32 &   0.23 &&    1.57 &  2.88 &  6.88 &  14.71 &  23.82 \\
\bottomrule
\end{tabular}
}
\begin{tabular}{llrrrrcrrrr}
\toprule
         & {} & \multicolumn{4}{c}{Accuracy} & \phantom{abc} & \multicolumn{4}{c}{Loss} \\
         &  & \multicolumn{4}{c}{\textbf{$\epsilon_{test}$}} && \multicolumn{4}{c}{\textbf{$\epsilon_{test}$}} \\
         \cmidrule{3-6} \cmidrule{8-11}
         & $\targetres$ &      0  &   3  &   5  &   10 && 0  &  3  &  5  &  10 \\
\cmidrule{2-6} \cmidrule{8-11}
\multirow{4}{*}{\imagenet} & 0.4 &     31.51 &  26.44 &  22.79 &  14.10 &&    3.71 &  4.14 &  4.55 &  5.89 \\
         & 0.5 &     37.94 &  31.90 &  23.61 &  12.81 &&     3.14 &  3.71 &  4.31 &  6.01 \\
         & 0.6 &     43.58 &  31.86 &  26.77 &  12.16 &&    2.68 &  3.39 &  3.96 &  6.42 \\
         & 0.7 &     49.54 &  36.92 &  28.62 &   9.14 &&    2.28 &  3.02 &  3.84 &  7.16 \\
         & 0.8 & 56.61 &  41.86 &  28.84 &   5.92 && 1.88 &  2.68 &  3.72 &  8.53 \\
\bottomrule
\end{tabular}
    \caption{Robust accuracy (in $\%$) and loss for \ct~, \ch~and \imagenet~under BPDA $\ell_2$-attack at different resolution scales. Attacked models are fine-tuned with $\targetres=0.6$.}
    \label{tab:app_l2}
\end{table*}


\subsection{FAD}
\begin{figure}[htbp]
    \centering
    \includegraphics[width=0.47\textwidth]{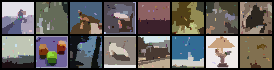}
    \includegraphics[width=0.47\textwidth]{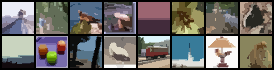}
    \caption{Bottom: Filtered Images. Top: Adversarial Images. Attacks remove important patches or add small ones.}
    \label{fig:attack_ex}
\end{figure}

\Cref{tab:app_l2} presents the results when attacking FAD with BPDA for \ct, \ch~and \imagenet, respectively. As the resolution increases, the attacker prevails the defense, and the robust accuracy decreases. The same holds for $\epsilon$. For values of $\epsilon$ bigger than the reported ones, accuracy quickly approaches zero. These behaviors are adherent to what one should expect. By reducing $\targetres$, fewer textures are present, and the attacker is less effective. As regards the attack strength, when $\epsilon$ increases, the attacker succeeds in creating new relatively small but strong patches or in removing patches that impact the shape (see~\Cref{fig:attack_ex}).

\begin{table*}[htbp]
\centering
\resizebox{0.95\textwidth}{!}{%
\begin{tabular}{rrrrrrcrrrrrr}
\toprule
& \multicolumn{5}{c}{$f_{nat}$} & \phantom{abc}& \multicolumn{5}{c}{$f_{aug}$}\\
\cmidrule{2-6} \cmidrule{8-12} 
        \textbf{Corruption} &  \textbf{Level 1} & \textbf{Level 2} & \textbf{Level 3} & \textbf{Level 4} & \textbf{Level 5} && \textbf{Level 1} & \textbf{Level 2} & \textbf{Level 3} & \textbf{Level 4} & \textbf{Level 5} \\
\cmidrule{2-6} \cmidrule{8-12} 
    $\mathtt{gaussian\_noise}$ &   78.8 &  59.45 &  40.54 &  33.52 &  28.78 &&  86.05 &  76.05 &  61.71 &  54.64 &  47.79\\
        $\mathtt{shot\_noise}$ &  86.81 &  77.78 &  55.88 &  47.29 &  35.05 &&  89.17 &  84.03 &  69.15 &  63.04 &  52.39 \\
     $\mathtt{impulse\_noise}$ &  85.39 &  74.64 &  64.77 &  42.31 &  26.59 &&  84.34 &  73.92 &  63.44 &  43.09 &  28.72\\
        $\mathtt{glass\_blur}$ &  53.34 &  53.26 &  59.27 &  41.63 &  47.08 &&  85.27 &  85.56 &  86.34 &  76.37 &  78.08\\
      $\mathtt{defocus\_blur}$ &  94.91 &  93.91 &  90.22 &  83.19 &  62.77 &&  92.95 &  92.92 &  92.59 &  91.29 &  88.35\\
      $\mathtt{motion\_blur}$ &  92.02 &  88.82 &  83.09 &  84.12 &   77.5 &&  91.14 &  90.37 &  88.51 &  88.71 &  85.29 \\
         $\mathtt{zoom\_blur}$ &  94.25 &  92.71 &  90.53 &  85.05 &  77.86 &&  92.87 &  93.09 &  92.87 &  92.23 &  91.28 \\
              $\mathtt{fog}$ &  94.77 &  93.98 &  93.16 &  91.07 &  79.07 &&  92.68 &  91.51 &  89.19 &  84.52 &  69.08 \\
             $\mathtt{frost}$ &  93.41 &  91.46 &  88.59 &  87.35 &  83.55 &&   92.0 &   91.0 &  89.03 &  88.44 &  85.49\\
              $\mathtt{snow}$ &  90.82 &  82.02 &  77.04 &  74.55 &  67.71 & &  91.7 &  88.82 &  83.03 &  78.65 &  77.42\\
          $\mathtt{contrast}$ &  94.73 &  93.19 &  91.73 &   87.8 &  56.06 &&  92.85 &  90.99 &  88.68 &  83.82 &  53.87 \\
        $\mathtt{brightness}$ &   94.9 &  94.72 &  94.33 &  93.95 &  92.55&&  92.99 &  92.81 &  92.49 &  91.91 &   90.4 \\
  $\mathtt{jpeg\_compression}$ &  87.27 &  82.08 &  79.91 &  77.25 &  73.96 &&  88.89 &  86.19 &  85.18 &  83.83 &   82.0\\
          $\mathtt{pixelate}$ &  92.24 &  87.91 &  82.65 &  64.62 &   44.4 &&  91.86 &  91.49 &  90.82 &  89.94 &  86.69\\
 $\mathtt{elastic\_transform}$ &  91.07 &   90.7 &  87.57 &  81.37 &  76.54 &&  90.24 &  90.84 &   90.9 &  88.64 &  85.07\\
\bottomrule
\end{tabular}
}
\caption{\ct. Accuracy (in $\%$) under data corruptions and different levels of severity. (Left) Natural model. (Right) Augmented model.}
\label{tab:corr_nat_aug}
\end{table*}
\subsection{Data Augmentation with Filtered Images}
As IGE removes textures, we may expect more robustness to corruptions. To test this hypothesis, we trained a ResNet-18 model $f_{aug}$ on CIFAR-10 by augmenting the training set of natural images $\mathbf{x}$ with their corresponding filtered counterparts $\mathbf{x}^{r^*}$ ($r^*=0.6$). The resulting augmented dataset is denoted $\D_{aug} = \{\D, \D^{r^*}\}$. The test set used for the evaluation is the original one (i.e., formed only by natural images).

As we may expect, the accuracy of $f_{aug}$ drops in relation to that of $f_{nat}$ (from roughly $95\%$ to $93.3\%$). 
However, if we look at the response of the two models under data corruption, $f_{aug}$ is almost always more accurate than $f_{nat}$. In some cases, e.g., glass blur, the gap is remarkable. The results are reported in~\Cref{tab:corr_nat_aug}. 
The code for applying data corruptions has been adapted from~\cite{salman2020unadversarial}.

\section{Discussion}
\label{sec:discussion}
In the following, we list the major limitations of IGE.
Regarding the implementation, although images are pre-processed using a GPU, it is still not ready for real-time applications. However, we pinpoint several improvements. 
The first is to increase computational performance by efficiently using the GPU architecture (e.g., threads and warping). The second direction regards the design of tailored merging rules to reduce the number of steps required.

We want to remark that we did not spend much effort tuning parameters and merging rules. Indeed, our main interest is to show that the model gains robustness by injecting the local stability of pixels.
We believe that the design of better merging rules will help improving accuracy and robustness.
Moreover, we expect further improvements by training models on a tailored resolution instead of using a unique model for all the resolutions.

\section{Conclusions}
\label{sec:conclusions}
This paper introduces a GPU-based image filtering framework, Image-Graph Extractor (IGE). Its outputs are the filtered image and the corresponding Image-Graph (IG). One of the several benefits of IGE is that it effectively enforces robustness to bounded perturbations by removing brittle textures from images.

Although the main focus is on the use of IGE for adversarial robustness, its flexible IG representation opens to several directions for future research. The first one concerns the use of the IG representation to cast image classification as a graph classification problem. This would allow to easily inject in the model user-defined data invariances (e.g., to the scale of objects). Another research direction consists in defining a specialized data augmentation strategy where nodes can be cleverly perturbed so as to remove common shortcut features such as the background color.

{\small
\bibliographystyle{ieee_fullname}
\bibliography{main.bib}
}
\clearpage

\appendix
\section{Implementation Details}
\label{sec:impl_details}
In order to satisfy parallel and efficient image processing we developed IGE in \texttt{CUDA/C++} and NVIDIA \texttt{Thrust}. The first piece of filtering is the extraction of connected-components. 
We use the GPU-based CCL (Connected Component Labeling) algorithm adapted from~\cite{playne2018new}.
The output of CCL is the $h\times w$ label matrix, where each location contains the identifier of the node/patch.
After running CCL, we run a CUDA-kernel called $\texttt{edge}$ which takes in input the label matrix $\texttt{edge <<< grid, block, block.x*block.y>>>}$ and outputs the graph (node, edge) in dataframe-like standard. 
That is, each connection consists of a row with source, destination and attributes.
Attributes are color distance, number of connected pixels, etc.
After the first phase,  the actual filtering procedure starts. 
At each step, the merge function is implemented in two phases: the first function has the objective to define the candidates for merging, while the second function actually merge nodes. 
After the merge, the filtered image is reconstructed with the function $\mathtt{avg\_color}$ that takes in input the label matrix and the original images and return the filtered images by averaging color for each node.
Moreover, the graph can be optionally saved on csv files and elaborated with the NVIDIA library \texttt{libcudf} (\url{https://github.com/rapidsai/cudf/)}. 

The python wrapper has been developed using \texttt{pybind11}. For future releases, we plan to develop a Torch extensions.

\section{IGE}
\label{app:ige}
With high-dimensional images, we have found beneficial to apply a de-noising pre-processing that consists of a partial removal of textures by the minimization of the following:
$$
\argmin_{\theta} L_{TV} (\theta) + L_{surf} (\theta) + \lambda \|x - \theta\|^2_2
$$
where $x$ is the raw image, $L_{TV}$ is the total variation loss and $L_{surf}$ is the surface loss. 
The objective of $L_{surf}$ loss is to minimize the number of changes in color pixel. In fact, total variation can be low even when the number of changes in pixels is very high. This happens for example when changes in color are minimal (e.g. 1 or 2 out of 255).
The change of pixels' color is defined with the step function $\mathbf{1}(\cdot)$, which is not differentiable.
We approximate the step function with the differentiable surrogate:
$$
\dfrac{1}{1 + e^{-{\alpha x}}} 
$$
where $\alpha \gg 1$.

\begin{figure*}[htbp]
    \centering
    \includegraphics[width=\fsize\textwidth]{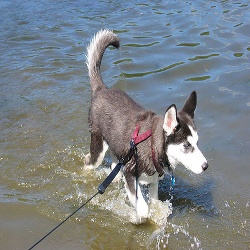}
    \includegraphics[width=\fsize\textwidth]{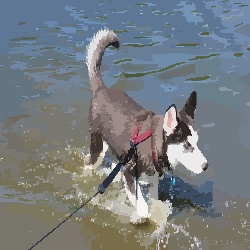}
    \includegraphics[width=\fsize\textwidth]{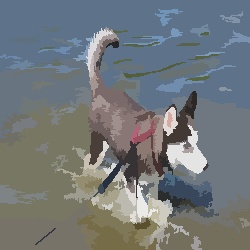}
    \includegraphics[width=\fsize\textwidth]{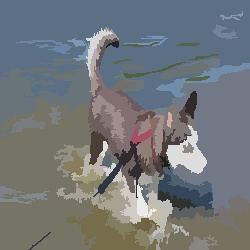}
    \includegraphics[width=\fsize\textwidth]{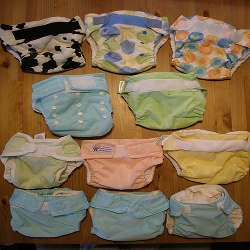}
    \includegraphics[width=\fsize\textwidth]{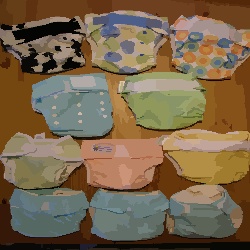}
    \includegraphics[width=\fsize\textwidth]{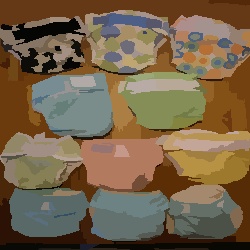}
    \includegraphics[width=\fsize\textwidth]{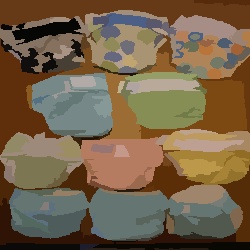}
    \includegraphics[width=\fsize\textwidth]{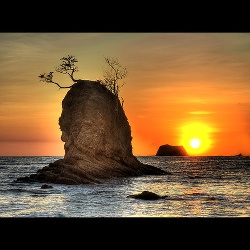}
    \includegraphics[width=\fsize\textwidth]{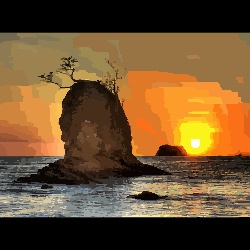}
    \includegraphics[width=\fsize\textwidth]{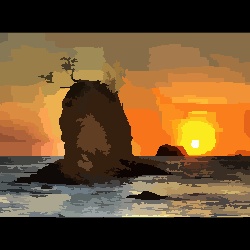}
    \includegraphics[width=\fsize\textwidth]{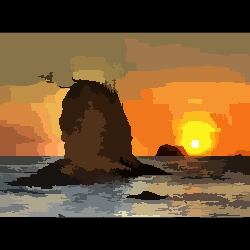}
    \includegraphics[width=\fsize\textwidth]{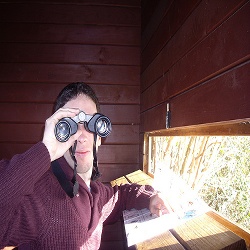}
    \includegraphics[width=\fsize\textwidth]{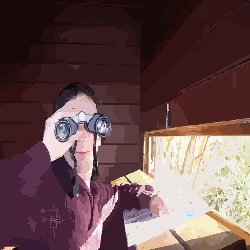}
    \includegraphics[width=\fsize\textwidth]{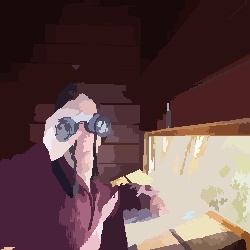}
    \includegraphics[width=\fsize\textwidth]{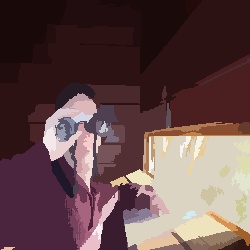}

    \caption{Examples of the effect of the IGE filtering when decreasing the target resolution. The values of the target resolutions are (from left to right): $1$, $0.7$, $0.45$, $0.35$.}
    \label{fig:app_res_ex}
\end{figure*}

\begin{figure*}
    \centering
    \includegraphics[width=0.4\textwidth]{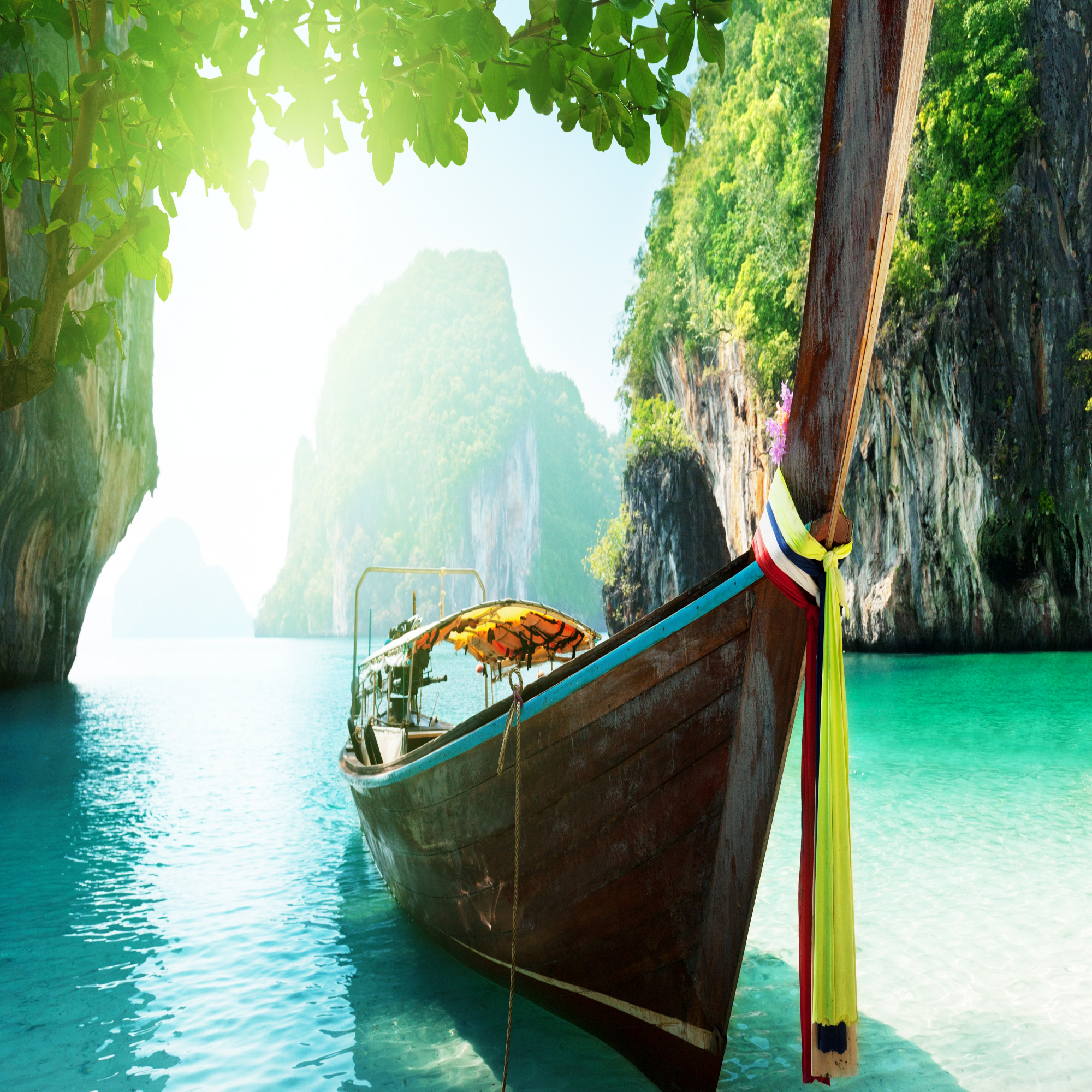}
      \includegraphics[width=0.4\textwidth]{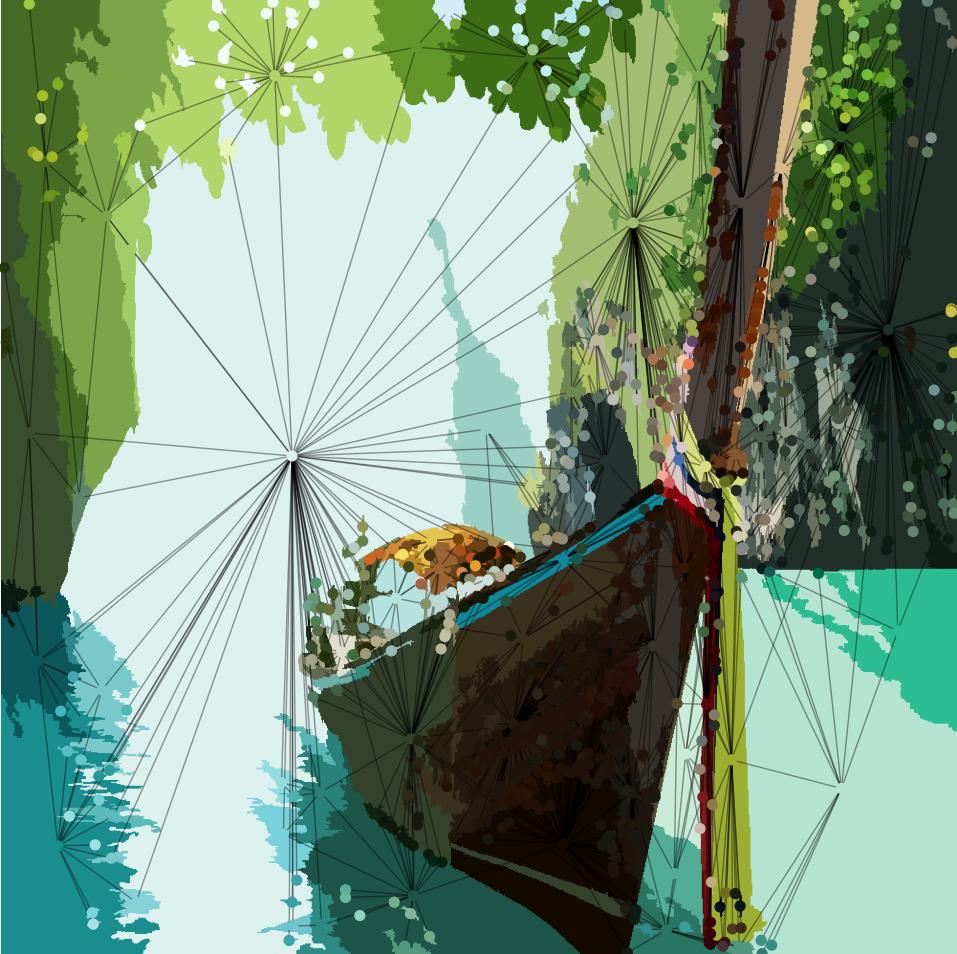}
    \includegraphics[width=0.5\textwidth]{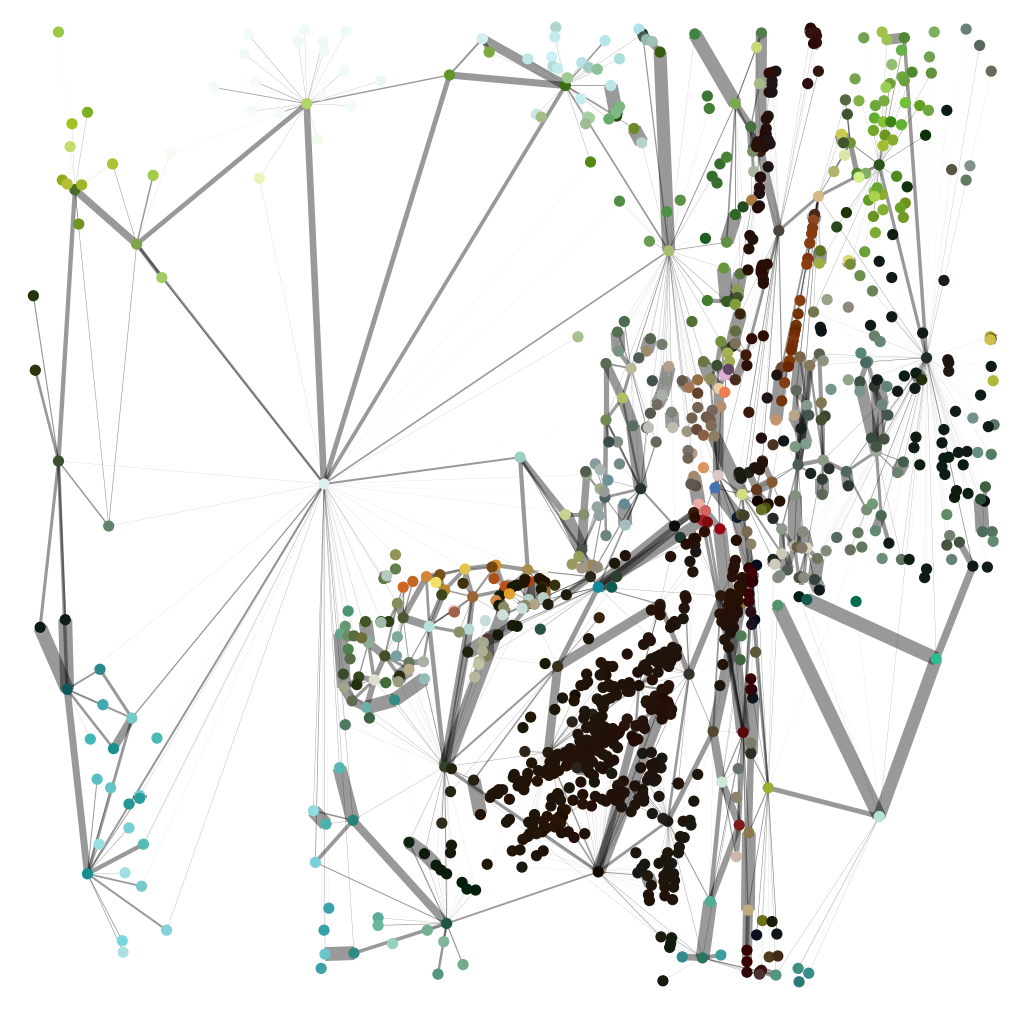}
    \caption{Example of Graph Image for a $1000\times 1000$ image. (Left) Original image. (Right) Graph over the image with $\targetres=0.5$. (Bottom) Image Graph. The width of edges is proportional to the fraction of perimeter shared by the source node with the target node.}
    \label{fig:app_graphs2}
\end{figure*}

\begin{figure*}
    \centering
      \includegraphics[width=0.4\textwidth]{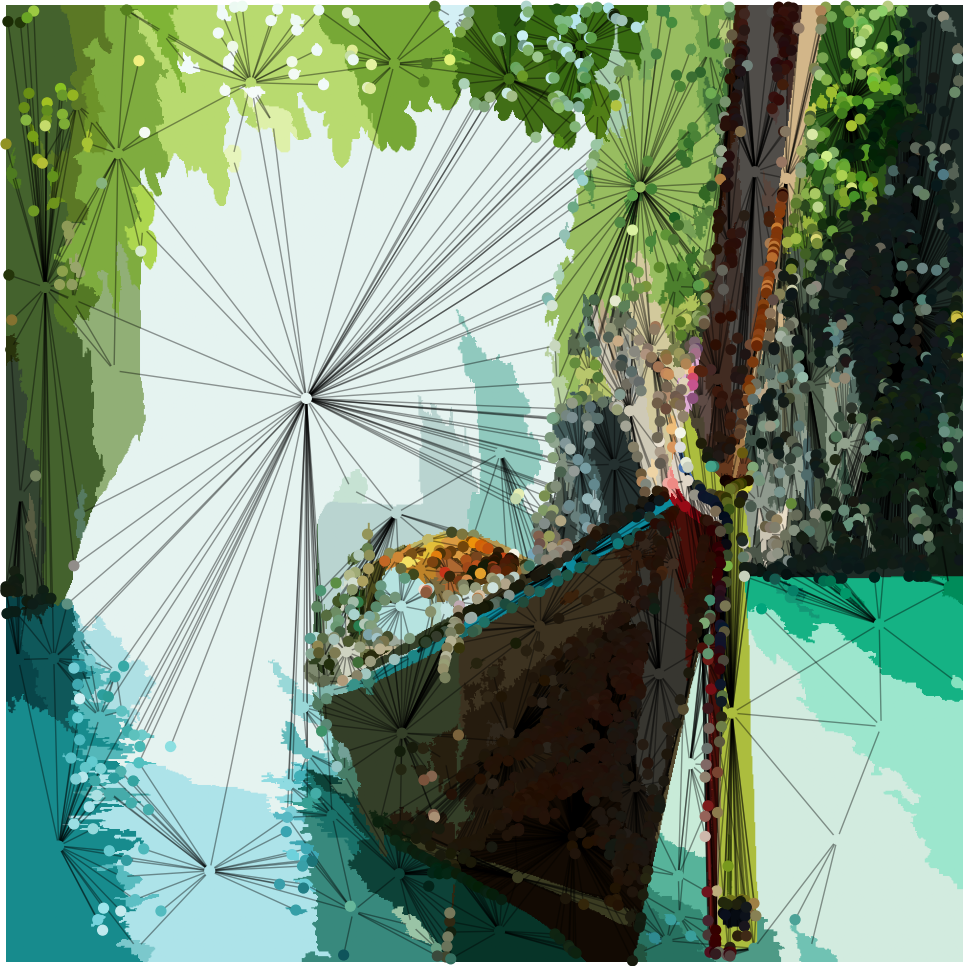}
    \includegraphics[width=0.4\textwidth]{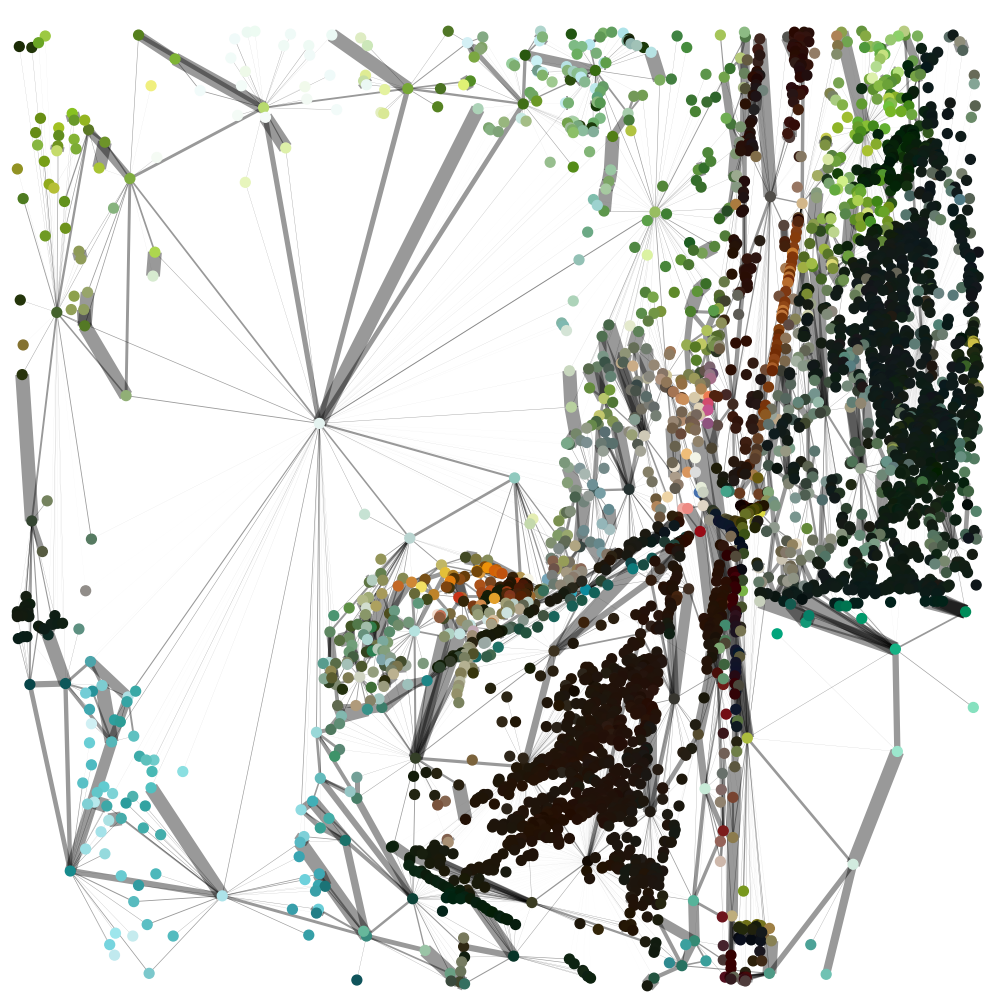}
    \caption{Example of Graph Image for a $1000\times 1000$ image. (Left) Original image. (Right) Graph over the image with $\targetres=0.6$. (Bottom) Image Graph. The width of edges is proportional to the fraction of perimeter shared by the source node with the target node.}
    \label{fig:app_graphs3}
\end{figure*}

In \Cref{fig:app_res_ex} are displayed filtered images at decreasing values of the target resolution $\targetres$.
Moreover, in \Cref{fig:app_graphs2,fig:app_graphs3} we present an example of computed graphs from different resolutions $\targetres$.

\section{FAD}
\label{app:fad}

Models are fine-tuned for $10$ epochs with learning rate $\eta=0.001$ and fixed resolution $\targetres=0.6$. For \imagenet, we sample $300$ filtered images per class from the training set. In \Cref{tab:app_l2} and \Cref{tab:app_linf} are reported the results when attacking FAD with BPDA with attacks bounded in $\ell_2$-norm and $\ell_\infty$-norm, respectively. As regards BPDA, we evaluate FAD with $50$ outer steps and $20$ inner steps. We did not take into account other white-box attacks (such as C\&W) since we consider PGD with BPDA to be the strongest attack we could evaluate against.


\begin{table*}[]
    \centering
    \resizebox{0.95\textwidth}{!}{%
    \begin{tabular}{llrrrrrcrrrrr}
\toprule
         & {} & \multicolumn{5}{c}{Accuracy} & \phantom{abc} & \multicolumn{5}{c}{Loss} \\
         &  & \multicolumn{5}{c}{\textbf{$\epsilon_{test}$}} && \multicolumn{5}{c}{\textbf{$\epsilon_{test}$}} \\
         \cmidrule{3-7} \cmidrule{9-13}
         & $\targetres$ & 0 &    1 &    2 &    3 &    4 &&      0 &   1 &   2 &    3 &    4 \\
\cmidrule{2-7} \cmidrule{9-13}
\multirow{5}{*}{\ct} & 0.4 &     78.45 &  70.15 &  55.37 &  35.09 &  21.16 &&    0.68 &  1.03 &  1.74 &   3.19 &   4.56 \\
         & 0.5 &     82.32 &  74.38 &  52.67 &  27.31 &  11.36 &&     0.54 &  0.85 &  1.91 &   4.17 &   6.48 \\
         & 0.6 &     85.06 &  73.18 &  49.80 &  18.26 &   4.39 &&    0.48 &  0.91 &  2.16 &   5.56 &   9.44 \\
         & 0.7 &     86.07 &  73.63 &  44.50 &   8.66 &   0.68 &&     0.44 &  0.91 &  2.68 &   8.14 &  13.95 \\
         & 0.8 &     89.55 &  74.48 &  33.40 &   1.82 &   0.00 &&     0.33 &  0.90 &  3.87 &  12.57 &  20.18 \\
\cmidrule{2-7} \cmidrule{9-13}
\multirow{5}{*}{\ch} & 0.4 &     41.08 &  34.99 &  25.91 &  19.04 &  15.01 &&     2.90 &  3.47 &  4.49 &   5.63 &   6.56 \\
         & 0.5 &     47.88 &  38.74 &  27.25 &  16.50 &  12.79 &&     2.47 &  3.17 &  4.59 &   6.35 &   7.58 \\
         & 0.6 &     52.60 &  41.24 &  25.52 &  13.93 &   7.49 &&     2.14 &  3.09 &  4.97 &   7.40 &   9.80 \\
         & 0.7 &     59.54 &  43.23 &  23.40 &   8.72 &   2.99 &&    1.80 &  3.03 &  5.45 &   9.54 &  13.82 \\
         & 0.8 &     63.74 &  45.31 &  19.17 &   3.32 &   0.23 &&    1.57 &  2.88 &  6.88 &  14.71 &  23.82 \\
\bottomrule
\end{tabular}
}
\begin{tabular}{llrrrrcrrrr}
\toprule
         & {} & \multicolumn{4}{c}{Accuracy} & \phantom{abc} & \multicolumn{4}{c}{Loss} \\
         &  & \multicolumn{4}{c}{\textbf{$\epsilon_{test}$}} && \multicolumn{4}{c}{\textbf{$\epsilon_{test}$}} \\
         \cmidrule{3-6} \cmidrule{8-11}
         & $\targetres$ &      0  &   3  &   5  &   10 && 0  &  3  &  5  &  10 \\
\cmidrule{2-6} \cmidrule{8-11}
\multirow{4}{*}{\imagenet} & 0.4 &     31.51 &  26.44 &  22.79 &  14.10 &&    3.71 &  4.14 &  4.55 &  5.89 \\
         & 0.5 &     37.94 &  31.90 &  23.61 &  12.81 &&     3.14 &  3.71 &  4.31 &  6.01 \\
         & 0.6 &     43.58 &  31.86 &  26.77 &  12.16 &&    2.68 &  3.39 &  3.96 &  6.42 \\
         & 0.7 &     49.54 &  36.92 &  28.62 &   9.14 &&    2.28 &  3.02 &  3.84 &  7.16 \\
         & 0.8 & 56.61 &  41.86 &  28.84 &   5.92 && 1.88 &  2.68 &  3.72 &  8.53 \\
\bottomrule
\end{tabular}
    \caption{Robust accuracy (in $\%$) and loss for \ct~, \ch~and \imagenet~under BPDA $\ell_2$-attack at different resolution scales. Attacked models are fine-tuned with $\targetres=0.6$.}
    \label{tab:app_l2}
\end{table*}

\begin{table*}[]
    \centering
    \resizebox{0.95\textwidth}{!}{%
   \begin{tabular}{llrrrrrcrrrrr}
\toprule
         & {} & \multicolumn{5}{c}{Accuracy} & \phantom{abc} & \multicolumn{5}{c}{Loss} \\
         &  & \multicolumn{5}{c}{\textbf{$\epsilon_{test}$}} && \multicolumn{5}{c}{\textbf{$\epsilon_{test}$}} \\ 
         \cmidrule{3-7} \cmidrule{9-13}
         & $\targetres$ & $0/255$ & $8/255$ & $16/255$ & $24/255$ & $32/255$ &&  $0/255$ & $8/255$ & $16/255$ & $24/255$ & $32/255$ \\
\cmidrule{2-7} \cmidrule{9-13}
\multirow{5}{*}{\ct} & 0.4 &     78.65 &  71.68 &  59.47 &  46.16 &  33.79 &&    0.68 &   0.93 &   1.45 &   2.16 &   3.15 \\
         & 0.5 &     83.79 &  74.02 &  61.78 &  43.72 &  26.79 &&     0.52 &   0.88 &   1.43 &   2.45 &   3.81 \\
         & 0.6 &     84.90 &  74.12 &  59.70 &  40.40 &  18.62 &&     0.47 &   0.88 &   1.53 &   2.80 &   5.19 \\
         & 0.7 &     86.91 &  74.28 &  56.51 &  28.74 &   2.12 &&     0.39 &   0.90 &   1.77 &   4.04 &  11.73 \\
         & 0.8 &     88.57 &  70.87 &  44.86 &   1.17 &   0.00 &&    0.37 &   1.06 &   2.72 &  13.60 &  25.87 \\
\cmidrule{2-7} \cmidrule{9-13}
\multirow{5}{*}{\ch} & 0.4 &     43.46 &  34.96 &  29.43 &  23.21 &  16.34 &&     2.79 &   3.41 &   4.07 &   4.84 &   5.78 \\
         & 0.5 &     48.57 &  41.34 &  30.24 &  23.18 &  15.85 &&     2.41 &   3.08 &   4.02 &   5.04 &   6.29 \\
         & 0.6 &     53.12 &  39.26 &  30.63 &  20.25 &  12.70 &&     2.13 &   3.15 &   4.13 &   5.57 &   7.38 \\
         & 0.7 &     59.90 &  43.36 &  29.07 &  17.58 &   5.89 &&     1.81 &   2.93 &   4.48 &   6.60 &  11.39 \\
         & 0.8 &     63.22 &  43.95 &  26.43 &   5.08 &   0.33 &&     1.55 &   3.03 &   5.38 &  13.80 &  28.71 \\
\cmidrule{2-7} \cmidrule{9-13}
\multirow{5}{*}{\imagenet} & 0.4 &     31.36 &  22.09 &  10.97 &   5.37 &   2.75 &&    3.71 &   4.68 &   6.85 &   9.14 &  11.65 \\
         & 0.5 &     37.67 &  24.29 &  10.60 &   4.53 &   2.21 &&     3.15 &   4.48 &   7.14 &  10.16 &  13.50 \\
         & 0.6 &     43.36 &  25.10 &  10.46 &   4.38 &   1.22 &&     2.68 &   4.29 &   7.50 &  10.99 &  15.46 \\
         & 0.7 &     49.65 &  26.83 &   8.73 &   2.55 &   0.39 &&    2.28 &   4.20 &   7.95 &  12.68 &  23.46 \\
         & 0.8 &     56.34 &  25.95 &   5.75 &   0.39 &   0.00 &&    1.90 &   4.37 &   9.71 &  22.82 &  57.72 \\
\bottomrule
\end{tabular}
}
    \caption{Robust accuracy (in $\%$) and loss for \ct~, \ch~and \imagenet~under BPDA $\ell_\infty$-attack at different resolution scales. Attacked models are fine-tuned with $\targetres=0.6$.}
    \label{tab:app_linf}
\end{table*}

\end{document}